\documentclass[runningheads]{llncs}

\usepackage{eccv}

\usepackage{eccvabbrv}

\usepackage{graphicx}
\usepackage{booktabs}

\usepackage{hyperref}

\usepackage{orcidlink}
\usepackage{multirow}

\begin{document}

\title{TripVVT: A Large-Scale Triplet Dataset and a Coarse-Mask Baseline for In-the-Wild Video Virtual Try-On} 

\titlerunning{TripVVT}

\author{Dingbao Shao\inst{1}$^\ast$ \and
Song Wu\inst{2}$^\ast$ \and
Shenyi Wang\inst{1} \and
Ye Wang\inst{3} \and
Ziheng Tang\inst{1} \and
Fei Liu\inst{4} \and
Jiang Lin\inst{1} \and
Xinyu Chen\inst{1} \and
Qian Wang\inst{2} \and
Ying Tai\inst{1} \and
Jian Yang\inst{1} \and
Zili Yi\inst{1}$^\dagger$}

\authorrunning{D. Shao et al.}

\institute{Nanjing University, China \and
JIUTIAN Research, CMCC, China \and
Jilin University, China \and
ByteDance Inc., China\\[2pt]
$^\ast$ Equal contribution. $^\dagger$ Corresponding author.}

\maketitle

\begin{center}
    \centering
    \vspace{-1.0em}
    \includegraphics[width=0.90\textwidth]{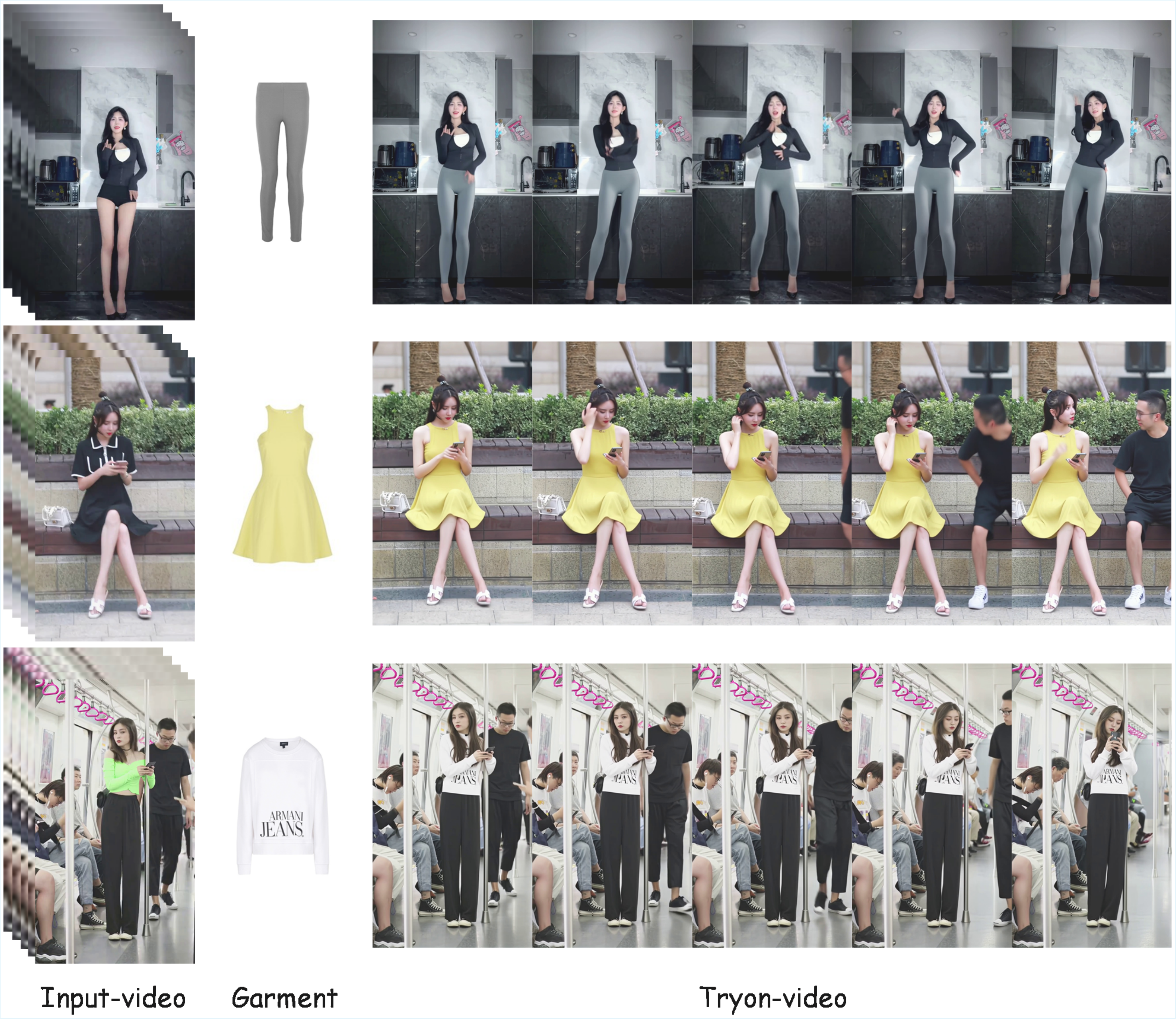}
    \captionof{figure}{Our method leverages the large-scale, diverse TripVVT-10K dataset and a coarse human-mask paradigm to produce realistic and temporally coherent try-on results under challenging real-world conditions. The top row illustrates dynamic motion in low-light scenes, the middle row shows seated poses that differ from the common standing cases, and the bottom row demonstrates performance in crowded public environments such as subway interiors, with complex backgrounds and multiple people in view. To support the research community, we have open-sourced the TripVVT-10K dataset and the associated benchmark.}
    \label{fig:teaser}
    \vspace{8pt}
\end{center}

\begin{abstract}
Due to the scarcity of large-scale in-the-wild triplet data and the improper use of masks, the performance of video virtual try-on models remains limited. In this paper, we first introduce \textbf{TripVVT-10K}, the largest and most diverse in-the-wild triplet dataset to date, providing explicit video-level cross-garment supervision that existing video datasets lack. Built upon this resource, we develop \textbf{TripVVT}, a Diffusion Transformer--based framework that replaces fragile garment masks with a simple, stable human-mask prior, enabling reliable background preservation while remaining robust to real-world motion, occlusion, and cluttered scenes. To support comprehensive evaluation, we further establish \textbf{TripVVT-Bench}, a 100-case benchmark covering diverse garments, complex environments, and multi-person scenarios, with metrics spanning video quality, try-on fidelity, background consistency, and temporal coherence. Compared to state-of-the-art academic and commercial systems, TripVVT achieves superior video quality and garment fidelity while markedly improving generalization to challenging in-the-wild videos. We publicly release the dataset and benchmark, which we believe provide a solid foundation for advancing controllable, realistic, and temporally stable video virtual try-on.

\vspace{2pt}
\noindent\textbf{Project page:} \url{https://shaodingbao.github.io/TripVVT/}
\end{abstract}

\section{Introduction}
\label{sec:intro}

Video Virtual Try-on (VVT) \cite{fang2024vivid, chong2025catv2ton, li2025magictryon} focuses on seamlessly replacing a person's clothing in video sequences with a given garment reference, while preserving key attributes such as the individual's identity, pose, motion, and background. It has gained significant attention due to its creative potential across various domains, including e-commerce, digital human creation, and the metaverse.

Current research in video virtual try-on can be broadly categorized into two paradigms: mask-based and mask-free approaches. Mask-based methods \cite{DBLP:conf/iccv/DongLSWC019,zhong2021mv,jiang2022clothformer,fang2024vivid,chong2025catv2ton,li2025magictryon}, due to the lack of suitable paired data, primarily focus on reconstructing region-level content from the original video, referencing garments directly from the input video itself. However, this reliance on high-quality masks limits their ability to generalize effectively to diverse garment references.
To overcome the dependency on masks, mask-free methods~\cite{chang2024pemf,zhang2025boow,wan2025mf} aim to directly generate the try-on results conditioned on the input video sequence and the reference garment, without requiring masks to explicitly specify the editing regions. However, such methods often inadvertently alter other regions, making it challenging to preserve the integrity of the background.  Moreover, most existing methods address virtual try-on in indoor video scenarios, but perform poorly on real, diverse, and complex in-the-wild video scenarios. These challenges significantly hinder the development and practical applications of video virtual try-on.

To address these challenges, we first introduce \textbf{TripVVT-10K}, the first large-scale and high-quality in-the-wild \textbf{Trip}let video dataset for \textbf{V}irtual \textbf{V}ideo \textbf{T}ry-on. \textbf{TripVVT-10K} comprises $10,031$ triplets of $\langle$original video, garment reference, try-on video$\rangle$, including 30 types of clothing categories from upper body, lower body, and full body, and covering diverse, real, and complex outdoor scenarios. We construct the dataset from an initial collection of web-crawled videos and filter out low-quality samples. The resulting high-quality subset forms the original video in triplet. Subsequently, we build an automated and efficient data synthesis pipeline based on Wan-Animate~\cite{wang2025omnitalkerrealtimetextdriventalking}. Specifically, we first use Nano Banana~\cite{comanici2025gemini} to create a high-quality swapped initial frame. We get the final try-on video via Wan-Animate under the guidance of the swapped frame and its original pose sequence. We then employ Nano Banana to extract the garment reference, which, together with original video and try-on video, constitutes the final triplet.

Leveraging TripVVT-10K, we introduce an end-to-end virtual try-on framework, TripVVT, based on a Diffusion Transformer (DiT) \cite{Peebles2022ScalableDM}. A novel hybrid paradigm is proposed to resolve the core trade-off between mask-based and mask-free methods. Its key design lies in using a simple human mask as the condition. This offers a stable spatial prior that confines edits to the human body, thereby robustly avoiding the over-constraints of garment masks and preserving the background. The model is trained using the try-on videos as input, with the original videos serving as the ground truth in a reverse design that facilitates robust supervision.

In addition, to support comprehensive evaluation and foster advancements, we introduce \textbf{TripVVT-Bench}, a dedicated benchmark tailored for in-the-wild scenarios. TripVVT-Bench incorporates a multi-dimensional evaluation framework across video quality, clothing fidelity, background preservation, and temporal consistency. We hope that TripVVT-Bench will serve as a standard benchmark for evaluating in-the-wild video virtual try-on and drive progress in this field.

\textbf{Our main contributions are as follows:}
\begin{itemize}
    \item We construct \textbf{TripVVT-10K}, the first large-scale, high-resolution in-the-wild \emph{video triplet} dataset for virtual try-on, enabling fully supervised training and evaluation under real-world conditions.
    \item We propose \textbf{TripVVT}, a Diffusion Transformer-based framework that harmoniously reconciles the mask-based and mask-free paradigms. By utilizing a coarse human mask as a spatial prior, it achieves state-of-the-art performance on both studio-style and challenging in-the-wild videos.
    \item We introduce \textbf{TripVVT-Bench}, the first comprehensive benchmark for video virtual try-on in real-world scenarios, providing a standardized and multi-dimensional evaluation platform covering video quality, try-on fidelity, background preservation, and temporal consistency.

\end{itemize}

\begin{table}[t]
\centering
\caption{Comparison of virtual try-on datasets. ``Pair'' denotes model--garment paired data; ``Pseudo'' denotes unpaired/pseudo paired image data; ``Triplet'' denotes video triplets of the same subject across garments.}
\label{tab:dataset_compare}
\small
\setlength{\tabcolsep}{3.5pt}
\resizebox{\textwidth}{!}{
\begin{tabular}{l c c c c c c c c}
\toprule
\multirow{2}{*}{\textbf{Dataset}} & \multirow{2}{*}{\textbf{Modality}} & \multirow{2}{*}{\textbf{Resolution}} & \multicolumn{2}{c}{\textbf{Scene}} & \multicolumn{2}{c}{\textbf{Person}} & \multirow{2}{*}{\textbf{Pairing}} & \multirow{2}{*}{\textbf{Number}} \\
\cmidrule(lr){4-5} \cmidrule(lr){6-7}
& & & Studio & Outdoor & Single & Multiple & & \\
\midrule
VITON-HD~\cite{choi2021viton}      & Image & 768$\times$1024 & \checkmark &            & \checkmark &        & Pair    & 11{,}647 \\
DressCode~\cite{morelli2022dress}  & Image & 768$\times$1024 & \checkmark &            & \checkmark &        & Pair    & 53{,}792 \\
StreetTryOn~\cite{cui2024street}   & Image & Var.            &            & \checkmark & \checkmark &        & Pseudo  & 14{,}453 \\
LAION-G.~\cite{guo2025any2anytryon}& Image & Var.            & \checkmark &            & \checkmark &        & Triplet & 60{,}000 \\
\midrule
VVT~\cite{DBLP:conf/iccv/DongLSWC019}               & Video & 192$\times$256  & \checkmark &            & \checkmark &        & Pair    & 791 \\
ViViD~\cite{fang2024vivid}         & Video & 624$\times$832  & \checkmark &            & \checkmark &        & Pair    & 9{,}700 \\
\midrule
\textbf{TripVVT-10K (ours)}        & Video & 720$\times$1280 &            & \checkmark & \checkmark & \checkmark & Triplet & \textbf{10,031} \\
\bottomrule
\end{tabular}
}
\end{table}

\section{Related Work}
\label{sec:related}

\subsection{Image-based Virtual Try-on}
Early image-based virtual try-on typically followed a ``warp--then--blend'' pipeline~\cite{morelli2023ladi,xie2023gp,yang2024d,zhang2023limb,zhu2023tryondiffusion}. With diffusion models, recent methods~\cite{choi2024improving,jiang2024fitdit,kim2024stableviton,li2023warpdiffusion,wang2024fldm,xu2025ootdiffusion} rely on human parsing or rule-based masks to preserve background and identity. While effective for single images, these fine-grained masks are sensitive to errors---tight masks truncate garment details, while loose masks overwrite background regions. Newer methods explore looser or optional spatial priors~\cite{jo2025up,kim2025promptdresser,guo2025any2anytryon,feng2025omnitry}, confirming that pixel-accurate garment masks are not essential. However, these approaches do not address the video-specific challenge of maintaining consistent spatial priors over time.

\subsection{Video Virtual Try-on}
Video try-on additionally requires temporal coherence and consistent motion following. Existing diffusion-based systems---such as ViViD~\cite{fang2024vivid}, CatV2TON~\cite{chong2025catv2ton}, and MagicTryOn~\cite{li2025magictryon}---extend image pipelines by incorporating temporal modeling. Similarly, Fashion-VDM~\cite{karras2024fashion} leverages advanced video diffusion architectures to significantly enhance temporal smoothness and garment details. However, these methods all fundamentally depend on precise garment masks or human parsing to explicitly define editing regions. These priors often become unreliable in real-world videos due to pose changes, occlusions, and lighting variations, leading to accumulated artifacts or background distortion. PEMF-VTO~\cite{chang2024pemf} further shows the difficulty of fully mask-free generation under limited pseudo-paired supervision. Collectively, prior works expose a fundamental tension: precise but fragile garment masks vs. flexible but unstable mask-free generation. This motivates our use of a coarse yet robust human-mask prior to balance spatial control and temporal stability.

\subsection{Datasets and Benchmarks}
Existing datasets for virtual try-on are predominantly image-based. VITON-HD~\cite{choi2021viton} and DressCode~\cite{morelli2022dress} provide paired person--garment images in studio environments, while Street TryOn~\cite{cui2024street} offers in-the-wild but unpaired single images. More recently, Any2AnyTryOn~\cite{guo2025any2anytryon} introduces the LAION-Garment dataset, which supplies image-level triplets of a person in one outfit, a garment reference, and the same person in another outfit, enabling supervised cross-garment transfer at the image level. On top of these datasets, VTBench~\cite{xiaobin2025vtbench} organizes a comprehensive benchmark for image-based try-on. However, all these resources are static and do not provide temporal information for video modeling.

For videos, public resources such as VVT~\cite{DBLP:conf/iccv/DongLSWC019} and the video portion of ViViD~\cite{fang2024vivid} are captured indoors with limited scene diversity and only pairwise supervision. Prior works have repeatedly highlighted the need for larger, more diverse, and temporally reliable video datasets~\cite{chong2025catv2ton,chang2024pemf,li2025realvvt}, yet, to our knowledge, there is still no publicly available in-the-wild video dataset with high-resolution triplets and an accompanying benchmark jointly assessing fidelity, background preservation, and temporal consistency. As comprehensively compared in Table~\ref{tab:dataset_compare}, TripVVT-10K and TripVVT-Bench are designed to fill this critical gap.

\section{Method}

\label{sec:third_section}

In this section, we first present an overview of the TripVVT-10K dataset in Sec.~\ref{Dataset Overview}. We then detail the data synthesis pipeline in Sec.~\ref{sec:dataset_curation}. Finally, we introduce the TripVVT in Sec.~\ref{Architecture}.
\begin{figure*}[t]
    \centering
    \includegraphics[width=\textwidth]{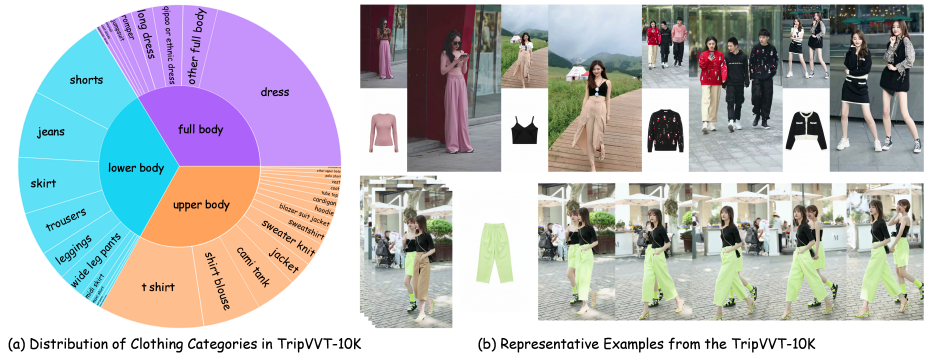}
    \caption{Overview of the TripVVT-10K Dataset. (a) Distribution of Clothing Categories: A sunburst chart visualizes the hierarchical composition of our dataset's wardrobe, spanning major categories like upper-body, lower-body, and full-body, each broken down into numerous fine-grained styles. (b) Representative Examples: A gallery of sample triplets showcases the dataset's diversity in capture conditions. These examples feature varied scenes, lighting, human dynamics, and the corresponding high-quality try-on results.}
    
    \label{fig:dataset}
\end{figure*}

\subsection{Dataset Overview}
\label{Dataset Overview}

TripVVT-10K is a large-scale video dataset comprising 10,031 triplets. It bridges the gap between controlled studio data and complex real-world environments. As shown in Fig.~\ref{fig:dataset}, the dataset demonstrates unprecedented diversity in both garment variety and capture conditions. This dual diversity is essential for developing robust models that can handle diverse clothing types while maintaining consistency in unconstrained settings. By systematically incorporating these variations, TripVVT-10K offers a valuable resource to advance virtual try-on technology toward practical real-world application.

\begin{figure*}[t]
    \centering
    \includegraphics[width=\textwidth]{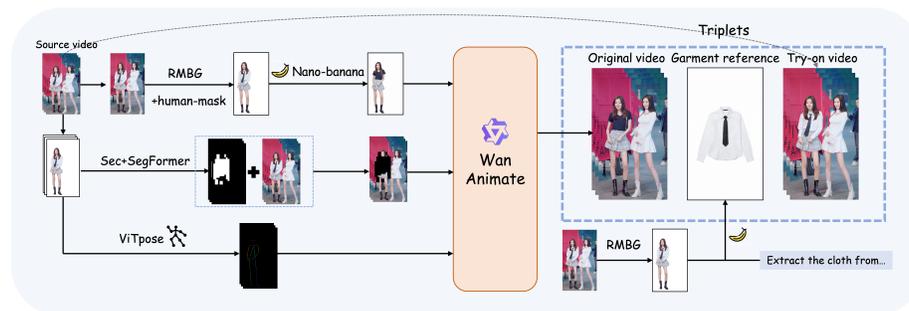}
    \caption{Overview of the TripVVT-10K data construction pipeline. We synthesize training triplets $\langle$original video, garment reference, try-on video$\rangle$ from in-the-wild source videos. (a) Original Video Synthesis: A garment-swapped anchor frame is first generated (via [11]); conditioned on this frame and extracted pose sequences, the source video is re-rendered through masked inpainting to create a synthesized "original" version. (b) Garment Reference: A canonical product image is generated by performing image-to-garment synthesis on the initial person crop. (c) Try-on Video: The raw source video serves as the ground-truth target. Our pipeline ensures spatial-temporal consistency across the generated triplets.}
    \label{fig:pipeline}
\end{figure*}

\subsection{Dataset Curation}
\label{sec:dataset_curation}

To support supervised learning for in-the-wild video virtual try-on, we construct \textbf{TripVVT-10K}, a large-scale triplet dataset consisting of $\langle$original video, garment reference, try-on video$\rangle$. Each triplet is derived from a real source video through an automated synthesis pipeline. Following a reverse design, we treat the synthesized garment-swapped video as the model input (original video), while the authentic source video is used as the ground-truth try-on video. This reverse-training paradigm ensures that the supervision signal always comes from real videos, avoiding the quality limitations of synthetic targets.

\paragraph{Stage 1: Source Video Collection and Preprocessing.}
We first crawl approximately 20K in-the-wild human videos from the web, covering diverse scenes, viewpoints, and garments. We normalize all videos to a fixed resolution, filter out low-quality or heavily compressed clips, and uniformly subsample each remaining video to 121 frames. The resulting high-quality subset serves as our \emph{source videos}.

To robustly track the target person, we combine Gemini-2.5-Flash~\cite{comanici2025gemini} and SAM2~\cite{ravi2024sam2} to obtain a temporally consistent \emph{human-mask video} for each source clip. These masks both provide clean person regions for downstream garment reconstruction and act as a key conditioning signal in our TripVVT model, serving as a simple yet effective spatial prior. Conditioning on human masks instead of fine-grained garment masks avoids the over-constrained behavior of traditional mask-based methods while still confining edits to the human body and preserving the background.

\paragraph{Stage 2: Triplet and Condition Generation.}
After preprocessing the source videos, we construct the three core components of each triplet $\langle$original video, garment reference, try-on video$\rangle$ together with auxiliary modalities required by our model. Following the reverse-training paradigm, we first synthesize the \emph{original video} (i.e., the model input), use the unmodified source video as the \emph{try-on video}, and then reconstruct the \emph{garment reference} from the source video.

To facilitate video synthesis, we additionally construct a \emph{garment-mask video} that is used only during generation. Concretely, we segment the garment region on the first frame using SegFormer~\cite{SegFormer} applied to the human-cropped person, and then propagate this garment mask across the entire sequence via SEC~\cite{zhang2025sec}.

\textit{First-frame editing.}
We begin by generating a swapped initial frame where the subject wears a new target garment as shown in Fig.~\ref{fig:pipeline}. A garment is randomly sampled from DressCode~\cite{morelli2022dress}, and Gemini-2.5-Flash~\cite{comanici2025gemini} is used to produce a textual description tightly bound to this specific garment instance. The subject portrait extracted from the first source frame is combined with this garment-specific description and fed into Nano Banana~\cite{comanici2025gemini} to synthesize a high-fidelity swapped first frame. By grounding each description on a concrete sampled garment and reusing it in Nano Banana, we ensure that the text is consistently matched to a specific garment, mitigating hallucination and repetition issues that arise when directly asking a language model to imagine garments.

\textit{Video synthesis via guided inpainting.}
Next, we synthesize a full garment-swapped video. We extract a pose sequence from the source video using ViTPose~\cite{xu2022vitpose} and prepare an inpainting-ready version of the video by masking out the original garment region with the propagated garment masks. Conditioned on the swapped first frame, the pose sequence, and the inpainting-ready source video, Wan-Animate performs guided inpainting to produce a temporally consistent garment-swapped sequence that preserves the original motion and background. This synthesized sequence is stored as the \emph{original video} in TripVVT-10K, while the unedited source video is directly used as the \emph{try-on video}.

\textit{Garment reference reconstruction.}
Finally, we construct the garment reference image corresponding to the clothing worn in the try-on video. We extract the subject from the first frame of the source video using the human mask and RMBG-2.0~\cite{BiRefNet} to remove complex background clutter and other people. The resulting clean foreground crop is then fed into Nano Banana~\cite{comanici2025gemini} to reconstruct a canonical product-style garment image, representing the clothing originally worn by the subject in a background-free, front-view style.

\paragraph{Stage 3: Quality Filtering.}
We apply a two-stage filtering procedure to ensure dataset quality. First, an automatic check based on Gemini-2.5-Flash evaluates the fidelity of the reconstructed garment reference and the consistency between the synthetic original video and the source video, considering attributes such as garment color, texture, and structure. Triplets with low garment fidelity or severe artifacts are discarded. Second, the remaining candidates are manually reviewed to remove cases with identity shifts, noticeable temporal flickering, or visible editing artifacts.

\paragraph{Augmentation.}
To further enrich the dataset beyond in-the-wild videos for training, we additionally incorporate supplemental data generated from existing public datasets. For image-based sources, we use \textbf{CatVTON}~\cite{chong2024catvtonconcatenationneedvirtual} to convert image garment-try-on pairs into triplets following the same $\langle$original video, garment\ reference, try\mbox{-}on video$\rangle$ structure. For video-based sources, we employ \textbf{MagicTryOn}~\cite{li2025magictryon} to synthesize video triplets in the same format. Since these synthetic image/video sources differ from TripVVT-10K in both content and generation characteristics, we design dedicated filtering strategies tailored to CatVTON and MagicTryOn outputs rather than reusing the TripVVT-10K pipeline. The detailed conversion and filtering procedures for these augmented samples are provided in the Supplementary Material. These samples are used solely as additional training data and are not included in the TripVVT-10K.

\paragraph{Auxiliary Modalities.}
Beyond the core triplet $\langle$original video, garment reference, try-on video$\rangle$, each TripVVT-10K sample is accompanied by several auxiliary conditions that are explicitly used by our TripVVT framework during training:  
(1) a \emph{human-mask video}, serving as the primary spatial prior and replacing traditional garment masks as the conditioning signal;  
(2) a \emph{pose video} estimated by DWPose~\cite{yang2023effective} from the original video, providing motion guidance;  
(3) a \emph{garment line map}, extracted from the garment reference using the AniLines tool~\cite{AniLines} to capture garment structure; and 
(4) a \emph{text prompt} describing the overall scene and the target person's outfit over the entire video, generated by Qwen-VL-Max~\cite{Qwen-VL}.
These auxiliary modalities are consistently generated for all triplets and tightly match the model design described in Sec.~\ref{Architecture}.

\subsection{Model Architecture}
\label{Architecture}
\begin{figure}
  \centering
  \includegraphics[width=\textwidth]{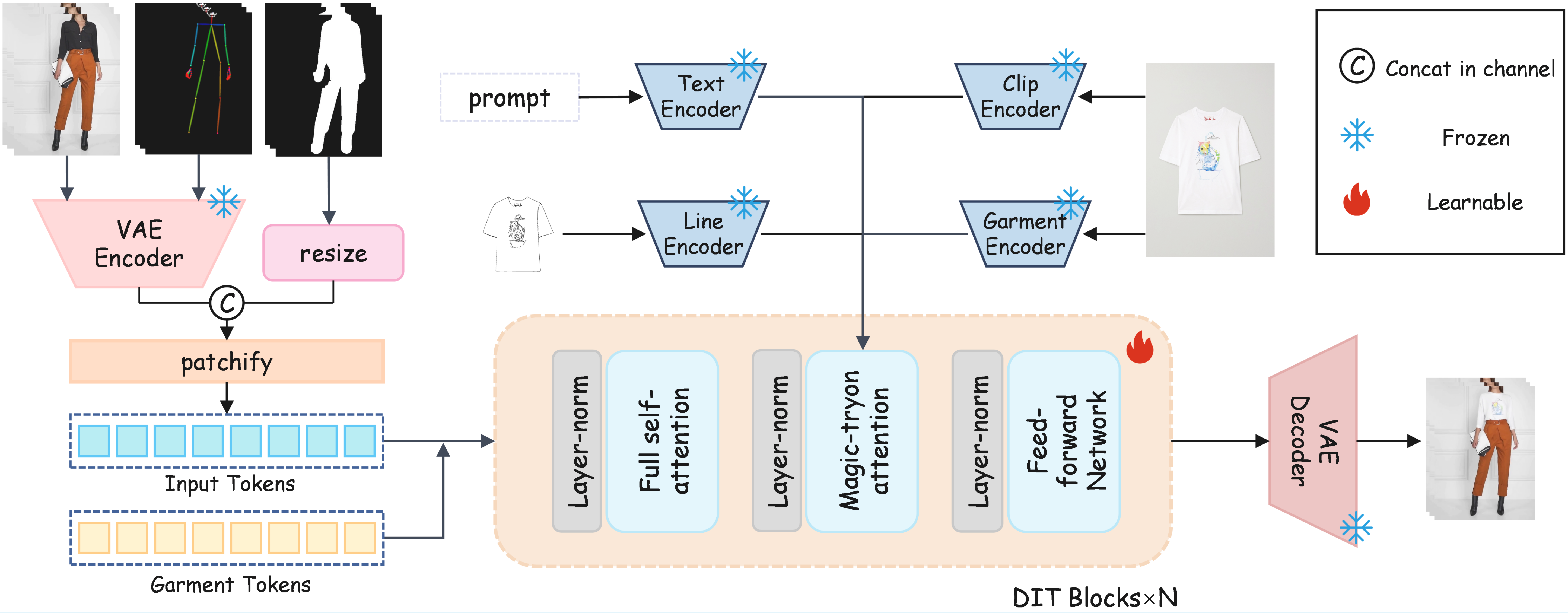}
  \caption{Architecture of TripVVT. The model takes the original video, pose and human mask as inputs, reusing the DiT-based backbone of MagicTryOn~\cite{li2025magictryon}, while garment and line encoders inject garment appearance and structural cues into the DiT blocks.}
  \label{fig:architecture}
\end{figure}

To effectively leverage the rich supervision provided by our TripVVT-10K dataset, we propose TripVVT, an end-to-end virtual try-on framework. As illustrated in Fig.~\ref{fig:architecture}, the model's architecture is specifically designed to capitalize on the dataset's structure. It builds upon the design of MagicTryOn~\cite{li2025magictryon} and is tailored to process the three synchronized inputs available in our dataset: the original video, the corresponding human-mask video, and the pose video of the target person.
We introduce these specific inputs to enhance control and robustness. The human mask serves two key functions: first, it preserves background consistency by strictly constraining the editing area to the subject, and second, it aids in localizing the person of interest. Compared to garment masks, human masks are also more robust and easier to obtain at high quality. The pose video, extracted via DWPose~\cite{yang2023effective}, is provided as an auxiliary condition to ensure the generated motion faithfully preserves the subject's original movements.
The core process begins with a VAE encoder extracting spatio-temporal features from the original video, which are then concatenated with the resized human-mask features. In parallel, the reference garment image is passed through a garment encoder to produce garment tokens. These fused features are fed into a DiT-based backbone. Finally, we condition the network on a global text prompt describing the entire scene (person, action, and environment) to serve as the textual control signal.

\section{Benchmark and Evaluation}
\label{sec:triptryon}

\subsection{TripVVT-Benchmark}
To address the limitations of existing benchmarks, which often rely on studio-only data and narrow evaluation metrics, we introduce TripVVT-Bench. Built upon the TripVVT-10K dataset, it consists of 100 high-quality and challenging triplets $\langle$original video, garment reference, try-on video$\rangle$ for in-the-wild evaluation, fully held out from the training set. The test set covers a diverse range of scenarios, including various garment types (upper, lower, and full-body), scenes with complex backgrounds, and both single- and multi-person videos. To ensure broad compatibility with different methods, we also provide auxiliary inputs such as garment masks and DensePose~\cite{DBLP:conf/cvpr/GulerNK18} information, generated using the methodology from CatV2TON~\cite{chong2025catv2ton}.
Our evaluation protocol for TripVVT-Bench is multi-faceted, assessing performance across four key dimensions. For video quality, we compute VFID (using both I3D~\cite{carreira2017quo} and ResNeXt~\cite{xie2017aggregatedresidualtransformationsdeep} backbones) and supplement it with the reference-based metrics SSIM~\cite{wang2004image} and LPIPS~\cite{zhang2018unreasonable}, applicable due to our ground-truth videos. For try-on fidelity, we use CLIP-I~\cite{radford2021learningtransferablevisualmodels} for semantic similarity and leverage Gemini-2.5-Flash~\cite{comanici2025gemini}(Gemini-SR) to automatically score the try-on success rate. Background consistency is evaluated via both pixel-level (BG-L1-Err) and perceptual (BG-DINO-Err\cite{simeoni2025dinov3}) differences in non-garment regions. Finally, for temporal consistency, we use CLIP-F~\cite{radford2021learningtransferablevisualmodels} to measure feature similarity between adjacent frames. Further details on these metric calculations are provided in the Supplementary Material.

\subsection{Implementation Details}
\label{sec: Implementation Details}

We adopt wan2.1-Fun-14B-control~\cite{WanFun14B} (fine-tuned from Wan2.1-I2V-14B~\cite{wan2025}) as our base model and train it with a three-stage progressive strategy that gradually increases data diversity and resolution. Stage 1 uses the TripVVT supplementary set for 25k steps at $512\times384$. Stage 2 adds TripVVT-10K and trains another 25k steps at mixed resolutions ($512\times384$ and $592\times334$). Stage 3 ups the resolution to $832\times624$ and $960\times544$ for 5.5k steps. All stages use 49-frame clips, batch size 2, AdamW~\cite{Loshchilov2017DecoupledWD} with a constant learning rate of $1\times10^{-5}$. Training runs on $8$ H100 GPUs.

\subsection{Experimental Settings}
We conduct a comprehensive evaluation on two distinct test sets to assess performance in both controlled and in-the-wild environments.
First, on the established ViViD-S test dataset~\cite{chong2025catv2ton}, we compare TripVVT against state-of-the-art academic methods, including ViViD~\cite{fang2024vivid}, CatV2TON~\cite{chong2025catv2ton}, and MagicTryOn~\cite{li2025magictryon}. We also include DreamVVT~\cite{zuo2025dreamvvt} in this comparison for publicly reported metrics. However, as DreamVVT is not open-source, we could not run our custom evaluation scripts on its outputs, so it is excluded from comparisons involving our proposed metrics on this test set.
Second, on our newly proposed TripVVT-Bench, which features challenging real-world scenarios and relies entirely on our comprehensive evaluation protocol, we benchmark TripVVT against the open-source academic models (ViViD, CatV2TON, MagicTryOn) and the commercial video editing tool Kling 1.6~\cite{kling_kuaishou_2025}.

\subsection{Quantitative Experiments}

\paragraph{Results on ViViD-S test.} As summarized in Tab.~\ref{tab:vivid-s}, TripVVT delivers competitive or superior results on most metrics on ViViD-S, with clear improvements in perceptual video quality and temporal consistency compared to prior methods. Thanks to the strong supervision of TripVVT-10K and its supplementary training sets, our model achieves strong try-on fidelity, while being slightly inferior to mask-based methods in background-related metrics and CLIP-based similarity. We attribute this gap in background consistency to the use of a coarse human mask instead of a fine-grained garment mask, which prioritizes robustness in real-world scenes over strict pixel-level alignment.

\begin{table*}
  \caption{Quantitative comparison with other methods on ViViD-S~\cite{chong2025catv2ton} test dataset. The best and second-best results are demonstrated in bold and underlined, respectively.}
  \label{tab:vivid-s}
  \centering
  \begin{tabular}{@{}l c c c c c c@{}} 
    \toprule
    Method & VFID$_I$ $\downarrow$ & VFID$_R$ $\downarrow$ & CLIP-I $\uparrow$ & CLIP-F $\uparrow$ & BG-L1 $\downarrow$ & BG-DINO $\downarrow$ \\ 
    \midrule
    ViViD~\cite{fang2024vivid}& 21.8032 & 0.8212 & \textbf{0.7270} & 0.9574 & 0.0571 & \textbf{0.0055} \\ 
    CatV2TON~\cite{chong2025catv2ton} & 19.5131 & 0.5283 & 0.7109 & 0.9304 & \underline{0.0507} & 0.0071 \\ 
    MagicTryOn~\cite{li2025magictryon} & 17.5710 & 0.5073 & 0.7072 & \underline{0.9647} & \textbf{0.0493} & \underline{0.0056} \\ 
    DreamVVT~\cite{zuo2025dreamvvt} & \underline{16.9468} & \underline{0.4285} & - & - & - & - \\ 
    Ours & \textbf{16.4398} & \textbf{0.3315} & \underline{0.7267} & \textbf{0.9670} & 0.0624 & 0.0059 \\ 
    \bottomrule
  \end{tabular}
\end{table*}

\begin{table*}[t]
  \caption{Quantitative comparison with other methods on TripVVT-Bench dataset. The best and second-best results are demonstrated in bold and underlined, respectively.}
  \label{tab:tripvvtbench}
  \centering
  \resizebox{\textwidth}{!}{
  \begin{tabular}{@{}l c c c c c c c c c@{}} 
    \toprule
    Method 
    & VFID$_I \downarrow$ 
    & VFID$_R \downarrow$ 
    & SSIM $\uparrow$
    & LPIPS $\downarrow$
    & CLIP-I $\uparrow$ 
    & CLIP-F $\uparrow$ 
    & BG-L1 $\downarrow$ 
    & BG-DINO $\downarrow$
    & Gemini-SR $\uparrow$\\
    \midrule
    ViViD~\cite{fang2024vivid}        & 26.7620 & 0.7083 & 0.8393 & 0.8393 & \textbf{0.7249} & 0.9349 & \underline{0.0691} & \underline{0.0053} & 9\%\\ 
    CatV2TON~\cite{chong2025catv2ton} & 34.2275 & 3.3266 & 0.7845 & 0.2130 & 0.6394 & 0.9212 & 0.1032 & 0.0109 & 4\% \\ 
    MagicTryOn~\cite{li2025magictryon}& 22.9238 & \underline{0.5502} & \textbf{0.8641} & \underline{0.1150} & 0.6934 & 0.9495 & \textbf{0.0515} & \textbf{0.0036} & 43\% \\ 
    Kling~\cite{kling_kuaishou_2025}   & \underline{22.4242} & 0.6289 & 0.6662 & 0.1353 & \underline{0.7148} & \underline{0.9504} &0.1662 & 0.0064 & \underline{78\%} \\ 
    Ours                              & \textbf{20.7245} & \textbf{0.3163} & \underline{0.8538} & \textbf{0.1053} & 0.7110 & \textbf{0.9606} & 0.0852 & 0.0059 & \textbf{91\%} \\ 
    \bottomrule
  \end{tabular}
  }
\end{table*}
\paragraph{Results on TripVVT-Bench.} On the more challenging TripVVT-Bench (Tab.~\ref{tab:tripvvtbench}), TripVVT clearly outperforms both academic and commercial competitors in video quality and temporal stability, and achieves competitive try-on fidelity, while also narrowing the gap to mask-based methods in background preservation. This highlights the advantage of training on diverse in-the-wild data: our model generalizes well where studio-trained baselines struggle. On reference-based metrics, TripVVT is slightly worse than MagicTryOn in SSIM, likely because the full-body human mask forces the model to reconstruct the entire person during try-on, making exact pixel-level alignment more difficult.

\begin{figure}[t]
  \centering
  \begin{subfigure}{0.48\linewidth}
    \centering
    \includegraphics[width=\linewidth]{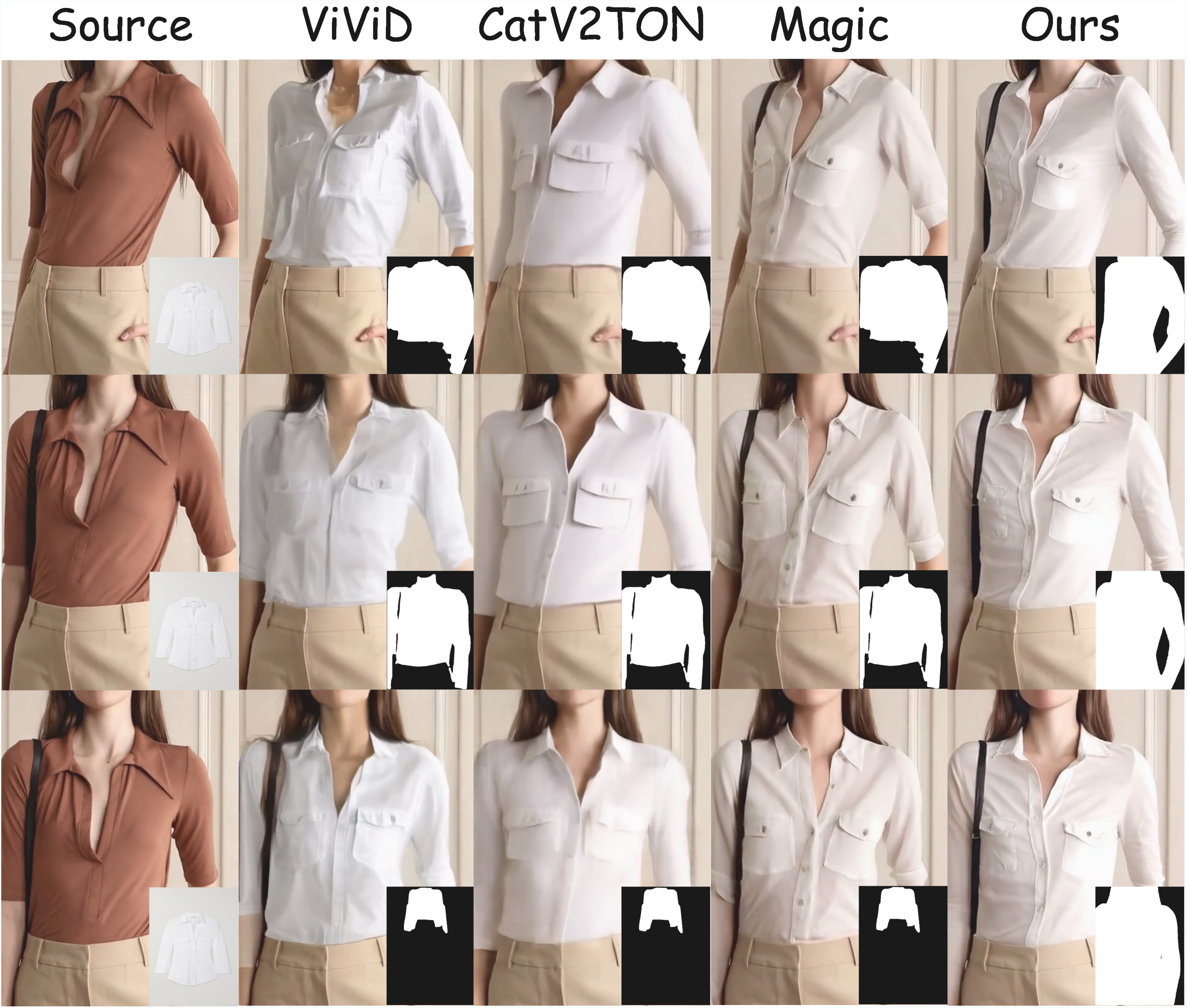}
    \caption{Qualitative comparison on the ViViD-S test dataset.}
    \label{fig:maskbad}
  \end{subfigure}
  \hfill
  \begin{subfigure}{0.5\linewidth}
    \centering
    \includegraphics[width=\linewidth]{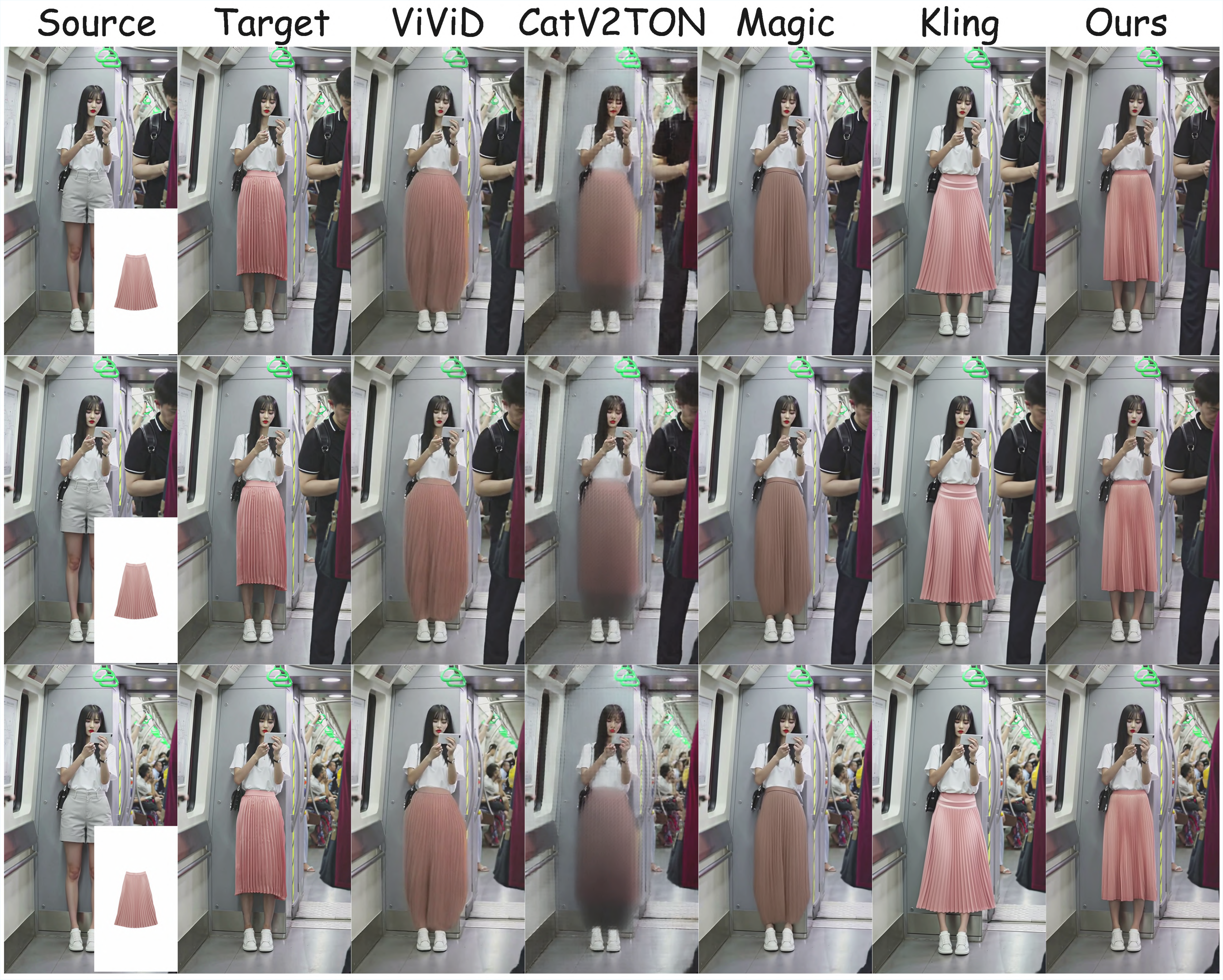}
    \caption{Qualitative comparison on the TripVVT-Bench.}
    \label{fig:tripexp}
  \end{subfigure}
  
  \caption{Overall qualitative comparisons. Left: results on ViViD-S; Right: results on our TripVVT-Bench.}
  \label{fig:combined_qualitative}
\end{figure}

\begin{table}[t]
  \centering
  \caption{Ablation study results on TripVVT-Bench.}
  \label{tab:ablation-table}
  \resizebox{\textwidth}{!}{
  \begin{tabular}{@{}l c c c c c c c c@{}} 
    \toprule
    Method & VFID$_I \downarrow$ & VFID$_R \downarrow$ & SSIM $\uparrow$ & LPIPS $\downarrow$ & CLIP-I $\uparrow$ & CLIP-F $\uparrow$ & BG-L1 $\downarrow$ & BG-DINO $\downarrow$ \\ 
    \midrule
    Base            & \textbf{20.7245} & \textbf{0.3163} & \textbf{0.8538} & \textbf{0.1053} & \underline{0.7110} & \textbf{0.9606} & \textbf{0.0852} & \textbf{0.0059} \\ 
    w/o TripVVT-10K & \underline{22.5105} & \underline{0.4569} & \underline{0.7605} & \underline{0.1182} & \textbf{0.7133} & 0.9483 & \underline{0.1552} & \textbf{0.0059} \\ 
    w/o Pose        & 33.6635 & 0.9949 & 0.5762 & 0.2744 & 0.6989 & 0.9540 & 0.3244 & 0.0112 \\ 
    w/o Human mask  & 33.2486 & 5.9749 & 0.6087 & 0.2570 & 0.6839 & \underline{0.9566} & 0.2942 & 0.0118 \\ 
    w Garment mask  & 28.5826 & 0.6946 & 0.5956 & 0.2451 & 0.7046 & 0.9466 & 0.3007 & 0.0102 \\ 
    \bottomrule
  \end{tabular}
  }
\end{table}

\subsection{Qualitative Experiments}
We provide qualitative comparisons in Fig.~\ref{fig:maskbad} and Fig.~\ref{fig:tripexp} to visually demonstrate the advantages of TripVVT.

\paragraph{Results on ViViD-S test.} Fig.~\ref{fig:maskbad} shows a single video sequence from the ViViD-S test dataset with three frames: the first frame has a poor garment mask, while the subsequent frames have relatively stable masks. In the first frame, where the garment mask is inaccurate, mask-based methods fail to reproduce the reference shirt's sleeve length and introduce noticeable artifacts. Thanks to the robust human-mask prior, TripVVT is the only method that correctly recovers the sleeve length while maintaining coherent edits. In the later frames with stable masks, our method still produces the most convincing results, with sharper details, more realistic fabric texture, and better alignment with the body.

\paragraph{Results on TripVVT-Bench.} Fig.~\ref{fig:tripexp} presents qualitative results on our challenging TripVVT-Bench. Since previous methods are mainly trained on indoor or studio-style data, they struggle to produce satisfactory try-on results in these in-the-wild scenarios, often showing severe distortions and inconsistent clothing appearance. The commercial video editing model Kling can successfully replace the outfit, but the generated garment texture and style differ noticeably from the reference image. In contrast, TripVVT better preserves human structure and produces more natural and realistic try-on effects, outperforming both open-source and commercial baselines. More qualitative results are provided in the Supplementary Material.

\subsection{User Study}
We conducted a user study on TripVVT-Bench to assess perceptual quality. For each triplet $\langle$original video, garment reference, try-on video$\rangle$, participants viewed anonymized outputs from all methods in random order and selected their top three preferences. They were asked to holistically balance video quality, try-on fidelity, background consistency, and temporal coherence; videos could be replayed and zoomed to reduce bias. We collected 2,500 votes from 50 participants. Following prior work, we report the proportion of first-place (Rank-1) votes as the primary metric. As shown in Table~\ref{tab:user_study}, our method is consistently preferred, achieving the highest user preference.

\begin{table}[t]
    \caption{User study results comparing different methods on TripVVT-Bench.}
    \label{tab:user_study}
    \resizebox{\linewidth}{!}{
    \setlength{\tabcolsep}{2.5pt}
    \begin{tabular}{cccccc}
        \hline
        Method & ViViD~\cite{fang2024vivid} & CatV2TON~\cite{chong2025catv2ton} & MagicTryOn~\cite{li2025magictryon} & Kling~\cite{kling_kuaishou_2025} & Ours \\
        \hline
        Rank 1 & 2.6\%  & 0.2\% & 12.2\% & 17.2\% & 67.6\% \\
        \hline
    \end{tabular}
    }
\end{table}

\subsection{Ablation Studies}
\label{ablation}
\begin{figure}
\centering
 
\includegraphics[width=1\linewidth]{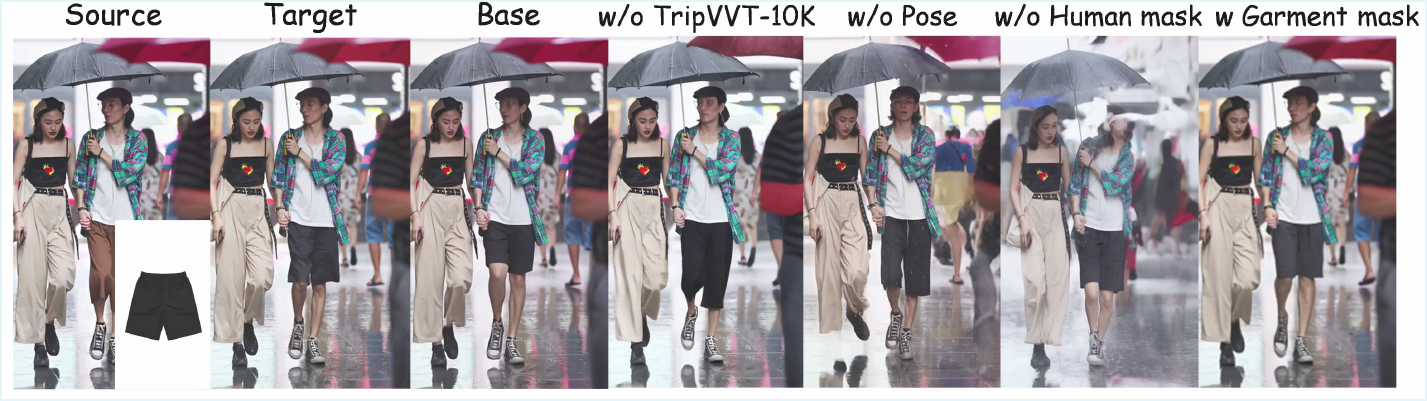}
\caption{Qualitative results of ablation studies on TripVVT-Bench.}
\label{fig:ablation}

\end{figure}

We conduct a series of ablation studies to validate the effectiveness of the key components in our approach as shown in Tab. \ref{tab:ablation-table} and Fig. \ref{fig:ablation}, respectively.

\paragraph{Effectiveness of the TripVVT-10K.} To demonstrate the crucial role of TripVVT-10K in enhancing generalization, we train a model variant exclusively on our supplementary dataset. As shown in Tab. \ref{tab:ablation-table}, this variant exhibits a significant performance degradation across all quantitative metrics compared to our full model. The qualitative comparison in Fig. \ref{fig:ablation} further reveals that without the rich, in-the-wild data from TripVVT-10K, the model fails to handle complex outdoor scenes, confirming the value of our curated dataset.

\paragraph{Effectiveness of the Human Mask.} To validate our human-mask prior, we evaluate two variants: removing the mask entirely, and replacing it with a garment mask (derived via CatV2TON~\cite{chong2025catv2ton}). As shown in Tab.~\ref{tab:ablation-table} and Fig.~\ref{fig:ablation}, both variants underperform our baseline. The mask-free variant suffers the most; without spatial constraints, the model struggles to localize edits, leading to severe boundary inconsistencies. Conversely, while the garment-mask variant improves upon the mask-free setting, it still introduces noticeable visual artifacts compared to our human-mask baseline. We attribute this performance gap to the inherent fragility and inaccuracy of garment masks in complex real-world scenarios.

\paragraph{Effectiveness of Pose Guidance.} Removing pose guidance has a minor impact on quantitative scores (Tab.~\ref{tab:ablation-table}), but qualitatively it leads to distorted poses and unnatural motion (Fig.~\ref{fig:ablation}), confirming its importance for maintaining motion fidelity.

\begin{figure}[!ht]
\centering
\includegraphics[width=0.55\linewidth]{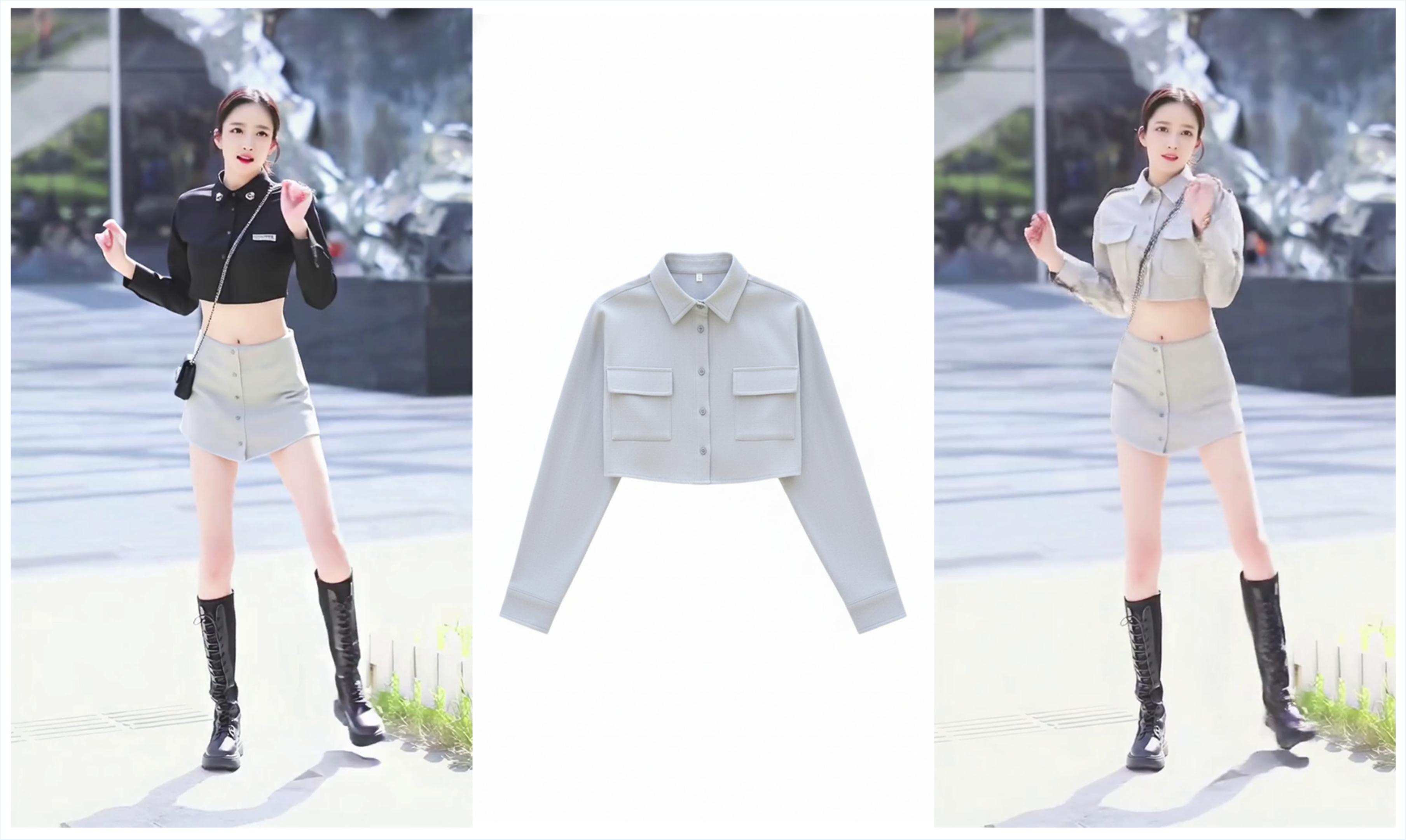}
\caption{A limitation example where the try-on model slightly alters non-target regions: the backpack disappears and minor changes occur on the skirt and boots during garment transfer.}
\label{fig:limited}
\end{figure}

\section{Limitations}
Our current pipeline relies solely on a human mask to localize the person to be edited, leaving the model to automatically infer which regions should undergo garment replacement. As a result, in some cases the model slightly alters clothing regions that are not intended for editing, as illustrated in Fig.~\ref{fig:limited}. This limitation suggests that relying on coarse spatial priors may be insufficient for strictly controlled garment transfer. Future work may explore new model architectures or training strategies that incorporate stronger structural guidance---such as explicit garment segmentation, region-level constraints, or attention supervision---to achieve more precise and disentangled try-on editing. 

\section{Conclusion}
\label{sec:conclusion}
We introduced TripVVT-10K, the first large-scale, high-resolution in-the-wild video triplet dataset, providing the explicit cross-garment supervision for training reliable video virtual try-on models. Built on this foundation, we developed TripVVT, a DiT-based framework that leverages a simple yet stable human-mask prior to balance spatial control with robustness in real-world scenes. To support standardized evaluation, we further proposed TripVVT-Bench, a 100-case benchmark for in-the-wild video virtual try-on.
Experiments on both ViViD-S and TripVVT-Bench demonstrate that TripVVT achieves strong video quality, garment fidelity, and temporal consistency, with clear advantages in challenging in-the-wild conditions where previous systems struggle. We hope that the dataset, benchmark, and method presented in this work provide a unified foundation for advancing controllable, realistic, and temporally stable video virtual try-on.

\bibliographystyle{splncs04}
\bibliography{main}
\end{document}